\documentclass{article}
\usepackage{ijcai24}

\usepackage{arydshln}
\usepackage{threeparttable}
\usepackage{CJK}
\usepackage{float}
\usepackage{makecell}
\usepackage{booktabs}
\usepackage{graphicx}
\usepackage{times}
\usepackage{soul}
\usepackage{url}
\usepackage[hidelinks]{hyperref}
\usepackage[all]{hypcap}
\usepackage[utf8]{inputenc}
\usepackage[small]{caption}
\usepackage{subcaption}
\usepackage{amsmath}
\usepackage{amsthm}
\usepackage{booktabs}
\usepackage{algorithm}
\usepackage{algorithmicx}
\usepackage{algpseudocode}
\usepackage[switch]{lineno}
\usepackage{color}
\usepackage{cleveref}
\usepackage{url}
\usepackage{multirow}
\usepackage[normalem]{ulem}
\useunder{\uline}{\ul}{}
\title{DiffStega: Towards Universal Training-Free Coverless Image Steganography\\with Diffusion Models}

\author{
Yiwei Yang$^1$\and
Zheyuan Liu$^1$\and
Jun Jia$^1$$^*$\and
Zhongpai Gao$^2$\and
Yunhao Li$^{1}$\and
Wei Sun$^{1}$\and\\
Xiaohong Liu$^1$\And
Guangtao Zhai$^{1}$\footnote{Corresponding authors.}\\
\affiliations
$^1$Shanghai Jiao Tong University\\
$^2$United Imaging Intelligence\\
\emails
\{evyoung, lzy233, jiajun0302\}@sjtu.edu.cn, 
gaozhongpai@gmail.com,\\
\{lyhsjtu, sunguwei, xiaohongliu, zhaiguangtao\}@sjtu.edu.cn\\
}
\begin{document}

\maketitle

\begin{abstract}
Traditional image steganography focuses on concealing one image within another, aiming to avoid steganalysis by unauthorized entities. Coverless image steganography (CIS) enhances imperceptibility by not using any cover image. Recent works have utilized text prompts as keys in CIS through diffusion models. However, this approach faces three challenges: invalidated when private prompt is guessed, crafting public prompts for semantic diversity, and the risk of prompt leakage during frequent transmission. To address these issues, we propose DiffStega, an innovative training-free diffusion-based CIS strategy for universal application. DiffStega uses a password-dependent reference image as an image prompt alongside the text, ensuring that only authorized parties can retrieve the hidden information. Furthermore, we develop Noise Flip technique to further secure the steganography against unauthorized decryption. To comprehensively assess our method across general CIS tasks, we create a dataset comprising various image steganography instances. Experiments indicate substantial improvements in our method over existing ones, particularly in aspects of versatility, password sensitivity, and recovery quality. Codes are available at \url{https://github.com/evtricks/DiffStega}.
\end{abstract}

\begin{figure}[ht]
  \centering
    \includegraphics[width=\linewidth]{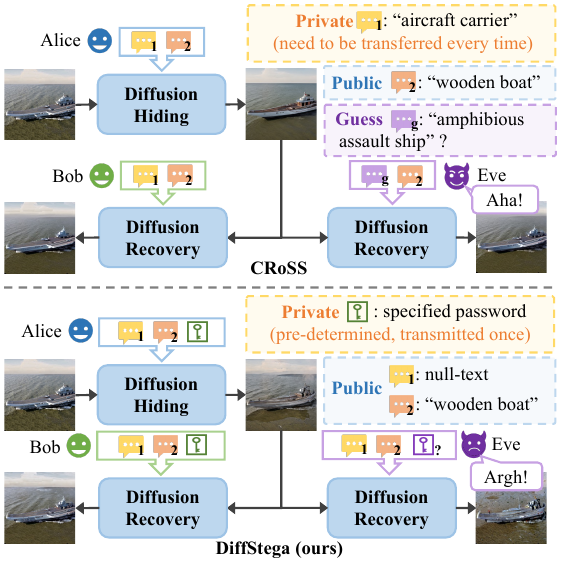}
  \caption{In this scenario, Alice represents a military organization that Eve regards as a target for espionage. Instead of using text prompt 1 as private key for diffusion-based CIS like previous work (CRoSS), DiffStega uses pre-determined password as private key, and null-text as prompt 1. DiffStega has no risk of text prompt leakage, and can encrypt the original image with arbitrary prompts.}
  \label{fig:teaser}
\end{figure}

\section{Introduction}
The internet revolution has significantly facilitated communication, yet posing challenges in securing messages transmitted over the Internet~\cite{mandal2022digital}. Steganography is a popular technique to hide information into the container in an imperceptible manner~\cite{yu2024cross}.
As a result, only trusted receivers are able to recover the information from the steganographic content. As a subset of this field, image steganography specializes in disguising the secret message as an image, offering a high degree of security and privacy. It has applications in diverse areas, including image compression~\cite{jafari2013increasing}, secure communication~\cite{duluta2017secure}, and cloud computing~\cite{alkhamese2019data}. Traditional cover-based image steganography schemes hide the secret message in a cover image by altering its statistical properties~\cite{meng2023review}. Once the cover image is leaked, the hidden message can be easily detected by steganalysis~\cite{karampidis2018review}. In contrast, coverless image steganography (CIS)~\cite{zhou2015coverless} aims to encode or map the secret message into a stego image rather than modifying a cover image. Thus, it has greater imperceptibility compared to cover-based techniques.

CRoSS~\cite{yu2024cross} has demonstrated the potential of using diffusion models for CIS. This approach consists of two stages: hiding and recovery. In the hiding stage, text prompt 1, serving as a private key, guides the Denoising Diffusion Implicit Model (DDIM)~\cite{song2020denoising} inversion process to convert an original image to an initial noisy latent code. Subsequently, text prompt 2, acting as a public key, directs the DDIM denoising process to produce a stego image from this latent code. However, the reliance on text prompts as private keys is vulnerable. As illustrated in~\Cref{fig:teaser}, if Eve knows about what Alice might encrypt through a background investigation, he could easily infer the correct private prompt. In practice, even a similar private prompt may be sufficient to breach the encryption. Furthermore, since the private key is contingent on the content of the original image, it must be transmitted for each use, creating a potential risk of leakage.

We posit that a general CIS task should offer substantial flexibility, meaning that the images intended for concealment could be encrypted into various contents or styles. To this end, we categorize the general CIS task into two distinct types: content-based steganography and style-based steganography. Content-based steganography alters the primary object categories within an image while preserving its inherent structure.
In contrast, style-based steganography achieves near-total imperceptibility by translating images into artistic renditions. This approach further obscures the detection and identification of hidden objects. Meanwhile, diffusion-based CIS should be able to produce stego images with arbitrary target text, especially general or similar texts.

To achieve general CIS, we introduce a novel diffusion-based pipeline.
Akin to traditional cryptography, our method hinges on a specific numerical password, predetermined and known only to trusted parties.
Since our password is independent of image contents, the password only needs to be transmitted once. Utilizing public text prompts and the prearranged password, trusted parties can generate a specific image to guide both the hiding stage and the recovery stage. To enhance the critical role of the password, we propose the Noise Flip technique, which encrypts the noisy latent code at the deepest step of the diffusion model while minimally altering its mean and variance. This innovation substantially complicates the task for potential attackers while simultaneously ensuring superior generation quality. To objectively assess our method's efficacy in general CIS tasks, we have curated a specialized dataset. This dataset comprises images alongside their target content, style, and analogous prompts. Our experimental results indicate significant advancements over existing methods, particularly in aspects of versatility, password sensitivity, and the effectiveness of recovery. 

Our contributions can be summarized as follows:
\begin{itemize}
\item We overcome the limitations of current diffusion-based CIS models by incorporating pre-determined passwords, leveraging existing models without further fine-tuning.
\item We introduce a novel CIS pipeline, which uniquely employs pre-determined passwords instead of text prompts alone to ensure security. This eliminates the need to transfer the private key each time the secret image is changed and is adaptable to any text prompt.
\item We create a specialized dataset tailored for general CIS tasks. Comprehensive experiments show the superior performance of ours in comparison to existing methods.
\end{itemize}

\section{Related Work}
\subsection{Image Steganography}
\paragraph{Cover-based Methods.}
Traditional cover-based methods are divided into spatial domain-based methods that directly modify the pixels of the cover image~\cite{yang2008adaptive,pevny2010using}, and transform domain-based methods that embed information into frequency domains~\cite{chen2007color,mckeon2007strange,valandar2017new}.
Deep learning-based methods use neural networks to hide information in cover images\cite{baluja,jia2022learning,jia2020rihoop,jia2022rivie}. SteganoGAN~\cite{steganogan} uses generative adversarial networks to optimize image  quality. HiNet~\cite{jing2021hinet} introduces invertible neural networks into steganography tasks. Changing the cover image leads to the usual drawbacks of cover-based methods with steganalysis~\cite{karampidis2018review}. Related topics include Image Forgery Detection~\cite{guo2023hierarchical} and invisible Watermarking~\cite{fu2024rawiw} have also been studied extensively~\cite{liu2022pscc,wu2023cheap,gao2015invisible}.

\paragraph{Coverless Methods.}
Coverless methods hide information without cover images. Early CIS refers to mapping the secret message into another image without modification~\cite{meng2023review}. Therefore, it fundamentally resists steganalysis and significantly improves security. 
Recently, CRoSS~\cite{yu2024cross} uses diffusion models and the inversion technique to achieve CIS. It uses prompts as private and public keys to translate the secret image into another, which is more controllable and robust with high quality. Diffusion-based CIS is promising with the powerful generation ability and rapid development of diffusion models.

\subsection{Diffusion Models}
Diffusion models are the newly emerged generative models, which synthesize images via progressively denoising Gaussian noise. Among them, DALLE-3~\cite{betker2023improving}, Imagen~\cite{saharia2022photorealistic} and Stable Diffusion~\cite{rombach2022high} have achieved the state-of-the-art results on many computer vision tasks.

\begin{figure*}[ht]
  \centering
    \includegraphics[width=\linewidth]{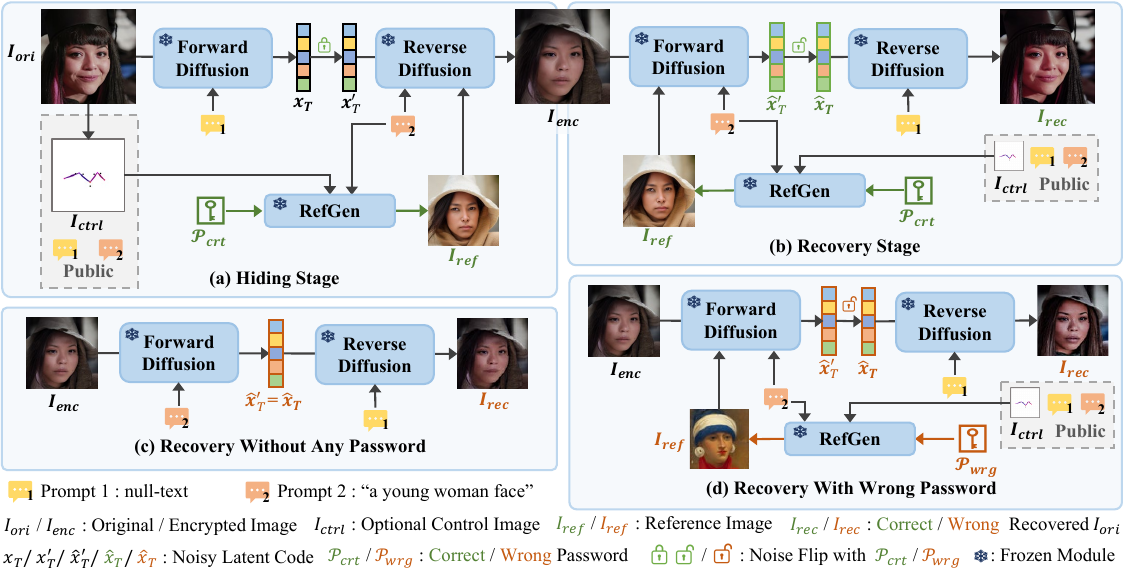}
  \caption{
    The pipeline of the DiffStega. 
    (a) We use text prompts and \textcolor[rgb]{0.33, 0.51, 0.2078}{${I}_{ref}$} generated by RefGen with password \textcolor[rgb]{0.33, 0.51, 0.2078}{$\mathcal{P}_{crt}$} to guide the diffusion process of hiding stage. The text prompts and the optional control image ${I}_{ctrl}$ (e.g. OpenPose bone image) is set public.
    (b) With public resources, authenticated parties could reproduce the same \textcolor[rgb]{0.33, 0.51, 0.2078}{${I}_{ref}$} with \textcolor[rgb]{0.33, 0.51, 0.2078}{$\mathcal{P}_{crt}$} to guide the diffusion process of recovery stage with text prompts. 
    (c) It illustrates the scenario where attackers attempt to directly recover the image without any password.
    (c) Wrong password \textcolor[rgb]{0.773, 0.353, 0.067}{$\mathcal{P}_{wrg}$} would result in wrong \textcolor[rgb]{0.773, 0.353, 0.067}{${I}_{ref}$}, which is distinct from the correct reference image, resulting in misleading the recovery diffusion process.
    \textcolor[rgb]{0.33, 0.51, 0.2078}{\textbf{Green}}\,/\,\textcolor[rgb]{0.773, 0.353, 0.067}{\textbf{Red}} denotes the \textcolor[rgb]{0.33, 0.51, 0.2078}{\textbf{correct}}\,/\,\textcolor[rgb]{0.773, 0.353, 0.067}{\textbf{wrong}} decrypted items. For brevity, we omit the encoder and decoder of VAE for latent diffusion models.}
  \label{fig:method}
\end{figure*}

\paragraph{Controlled Generation.}
Recent advances provide additional control for diffusion process.
ControlNet~\cite{zhang2023adding} and T2I-Adapter~\cite{mou2023t2i} leverage adapters to add conditional controls of semantic segmentation, OpenPose bone image, and Canny edge detection, etc.
In addition, BLIP-Diffusion~\cite{li2023blip} and IP-Adapter~\cite{ye2023ip} handle reference images as additional prompts to control the generation with image embeddings.

\paragraph{Diffusion Inversion.}
Diffusion inversion deterministically noise the image to the intermediate latent code along the path that the denoising would follow with the same conditioning~\cite{wallace2023edict}, widely used for image editing tasks. Among them, DDIM~\cite{song2020denoising} relies on local linearization assumptions, suffering errors with the actual process. To reduce inversion errors, Null-text Inversion~\cite{mokady2023null} introduces additional fine-tuning on optimizing the null embedding, and EDICT~\cite{wallace2023edict} uses coupled transformations without fine-tuning.

\section{Method}
\subsection{Overview}
Our method consists of two stages: hiding and recovery.
In hiding stage, shown in~\Cref{fig:method}a, we first generate a reference image $I_{ref}$ with given password $\mathcal{P}_{crt}$, prompt 2 and optional control image $I_{ctrl}$ (e.g. OpenPose bone image). Then we translate original secret image $I_{ori}$ into encrypted image $I_{enc}$ with the guidance of prompt 1, prompt 2, and $I_{ref}$. We set prompt 1, prompt 2 and $I_{ctrl}$ as public resources. In recovery stage, shown in~\Cref{fig:method}b, the secret image is recovered from $I_{enc}$ with correct password $\mathcal{P}_{crt}$ and public resources in a reverse procedure of hiding stage. If a malicious attacker attempts to directly recover the image without any passwords, or with a wrong password, a distinct recovery would be obtained, shown in~\Cref{fig:method}c and in~\Cref{fig:method}d respectively. To control the diffusion process in the two stages, we propose Reference Generator (RefGen) to generate $I_{ref}$ as image prompts, and Guidance Injection module as detailed in~\Cref{sec:refgen} and ~\Cref{sec:gim} respectively.

\subsubsection{Hiding Stage}
In hiding stage, DiffStega mainly uses two encryption methods. One is RefGen to generate $I_{ref}$ that guides the reverse diffusion process. Another is Noise Flip to ensure that the original image cannot be recovered with wrong passwords. We introduce the hiding stage as follows.

\paragraph{Preparation.} We first input private password $\mathcal{P}_{crt}$, public text prompt 2, and optional control image ${I}_{ctrl}$ (e.g. OpenPose bone image, semantic image) into RefGen, to generate a reference image $I_{ref}$. Note that we only set text prompts and ${I}_{ctrl}$ public. $\mathcal{P}_{crt}$ and $I_{ref}$ will not be published publicly.

\paragraph{Forward Diffusion.} Since we use lattent diffusions~\cite{rombach2022high}, $I_{ori}$ is first encoded by VAE into latent codes $x_{0}$. Then we use diffusion inversion to convert $x_{0}$ to initial noisy latent code $x_{T}$, where T is DDIM~\cite{song2020denoising} steps. Since we choose EDICT~\cite{wallace2023edict} as noising inversion method, our method is training-free. This process requires only null-text prompt 1 as guidance.

\paragraph{Noise Flip.} To improve the influence of $\mathcal{P}_{crt}$, we use it as random seed to deterministically flipping partial positions in $x_{T}$, resulting in a slightly different $x^\prime_{T}$. We denote this procedure as Noise Flip. The more positions flipped, the harder for attackers to recover the original image without the correct password. This Noise Flip process follows the formula:
\begin{equation}\nonumber
x^\prime_T =  x_T \odot {(1-M_{rand}(\mathcal{P}_{crt}, \eta))} - x_T \odot M_{rand}(\mathcal{P}_{crt}, \eta)
% \label{alg:noise_mix}
        % \resizebox{.89\linewidth}{!}{$
        %     \displaystyle
        %     x^\prime_T =  x_T \odot {(1-M_{rand}(\mathcal{P}_{crt}, \eta))} - x_T \odot M_{rand}(\mathcal{P}_{crt}, \eta)
        % $},
    % \small
    % \begin{aligned}
    %     x^\prime_T = - x_T \odot M_{rand}(\mathcal{P}_{crt}, \eta) + x_T \odot {(1-M_{rand}(\mathcal{P}_{crt}, \eta))},
    % \end{aligned}
\end{equation}

\noindent where $\odot$ denotes element-wise multiplication, and $M_{rand}$ is a random binary mask generated with deterministic random seed according to $\mathcal{P}_{crt}$. The proportion of 1 in $M_{rand}$ is controlled through the coefficient $\eta \sim [0,1]$. Small $\eta$ is sufficient for significant modification of $x_{T}$. After this procedure, the noisy latent code of step $T$ has strong dependence on $\mathcal{P}_{crt}$, while barely altering its mean and variance.

\paragraph{Reverse Diffusion.} In this process, we use Guidance Injection to treat $I_{ref}$ as image prompt to guide noise prediction together with text prompt 2. Then, the reverse diffusion process converts $x^\prime_{T}$ into $x^\prime_{0}$ with EDICT~\cite{wallace2023edict} denoising. Finally, $x^\prime_{0}$ is decoded into encrypted image $I_{enc}$ with VAE. With the aid of $I_{ref}$ and Noise Flip, $I_{enc}$ would be distinct from $I_{ori}$. Only authenticated users with $\mathcal{P}_{crt}$ could recover the original image. For brevity, we use one latent code to represent coupled latent pairs used in EDICT.

\subsubsection{Recovery Stage}
As illustrated in~\Cref{fig:method}, the recovery stage is the reverse process of the hiding stage in symmetry. With private $\mathcal{P}_{crt}$, authenticated parties could easily recover the original image, where attackers is hard to predict the correct numerical password. In this stage, we have encrypted image $I_{enc}$ and public resources consisting of prompt 1, prompt 2, and ${I}_{ctrl}$. The possible scenarios are that we have the correct password $\mathcal{P}_{crt}$, or the wrong password $\mathcal{P}_{wrg}$, or do not use any password at all. We will discuss the details of these scenarios as follows.

\paragraph{Recovery with Correct Password.} 
With $\mathcal{P}_{crt}$, public prompt 2 and ${I}_{ctrl}$, we first use RefGen to reproduce the correct $I_{ref}$, identical to that used in hiding stage. Then we use $I_{ref}$ and prompt 2 to guide noise prediction together via Guidance Injection. VAE encodes $I_{enc}$ into $\hat{x}^\prime_{0}$. The forward diffusion process convert $\hat{x}^\prime_{0}$ into $\hat{x}^\prime_{T}$ with EDICT noising inversion. Following the same procedure of Noise Flip in hiding stage, we could just flip back the previous reversed positions in $\hat{x}^\prime_{T}$, resulting in $\hat{x}_{T}\approx x_{T}$. Then we use reverse denoising process via EDICT to convert $\hat{x}_{T}$ to $\hat{x}_{0}$. After VAE decoder, we finally get correctly recovered image $I_{rec}$.

\paragraph{Recovery without Any Password.} For malicious attackers, the simplest way they would try is directly conducting recovery with only public text prompts, as shown in~\Cref{fig:method}c. It removes the guidance of $I_{ref}$ and Noise Flip from the correct recovery procedure. However, without the guidance of $I_{ref}$, there is a gap between $\hat{x}^\prime_{T}$ and $x^\prime_{T}$. Moreover, since flipped positions in $\hat{x}^\prime_{T}$ remain, $\hat{x}_{0}$ is distinct from ${x}_{0}$, leading to awfully wrong recovery. 

\paragraph{Recovery with Wrong Password.} As shown in~\Cref{fig:method}d, wrong password $\mathcal{P}_{wrg}$ would produce wrong $I_{ref}$, which is distinct from that used in hiding stage. It results in wrong $\hat{x}^\prime_{T}$ from $\hat{x}^\prime_{0}$ in the forward diffusion process. After Noise Flip with a wrong random seed, $\hat{x}_{T}$ would be further far different from the original $x_{T}$. Since prompt 1 is null-text, it is nearly impossible for the final $I_{rec}$ to present the content with the same semantics as the original image.

\subsection{Reference Generator}
\label{sec:refgen}
\begin{figure}[t]
  \centering
    \includegraphics[width=0.91\linewidth]{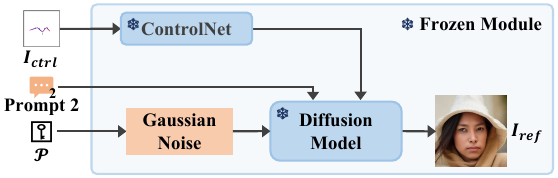}
  \caption{
    Details of Reference Generator. It first generates a deterministic initial Gaussian noise according to given password. $I_{ref}$ is generated from pretrained diffusion models with the guidance of $I_{ctrl}$ with ControlNet and prompt 2.}
  \label{fig:refgen}
\end{figure}

In this section, we will introduce the inside procedure of RefGen. It aims to produce deterministic images according to the password and $I_{ctrl}$. As shown in~\Cref{fig:refgen}, RefGen first generates a deterministic Gaussian noise with the random seed according to given password $\mathcal{P}$. The Gaussian noise serves as the initial noisy latent code of generation. We use ControlNet~\cite{zhang2023adding} to add additional control with $I_{ctrl}$ (e.g. OpenPose bone image) to diffusion models. Then the reference image $I_{ref}$ is obtained from pretrained diffusion models with the guidance of $I_{ctrl}$ and prompt 2.

\subsection{Guidance Injection}
\label{sec:gim}
This procedure aims to inject image features of $I_{ref}$ to the diffusion models.
% Latent diffusion models~\cite{rombach2022high} use CLIP~\cite{radford2021learning} to convert texts to embeddings, and then guide the diffusion process with the cross-attention layers of U-Net on the low dimensional latent code. 
In order to use images to guide the diffusion process in the same way as the text, we adapt IP-Adapter~\cite{ye2023ip} to inject image features into U-Net on the low dimensional latents. Furthermore, we notice that if we use optional control image $I_{ctrl}$ to force $I_{ref}$ have the similar structure with $I_{ori}$, the encrypted image $I_{enc}$ usually have the similar structure with the aid of Guidance Injection.

\begin{figure*}[th]
  \centering
    \includegraphics[width=\linewidth]{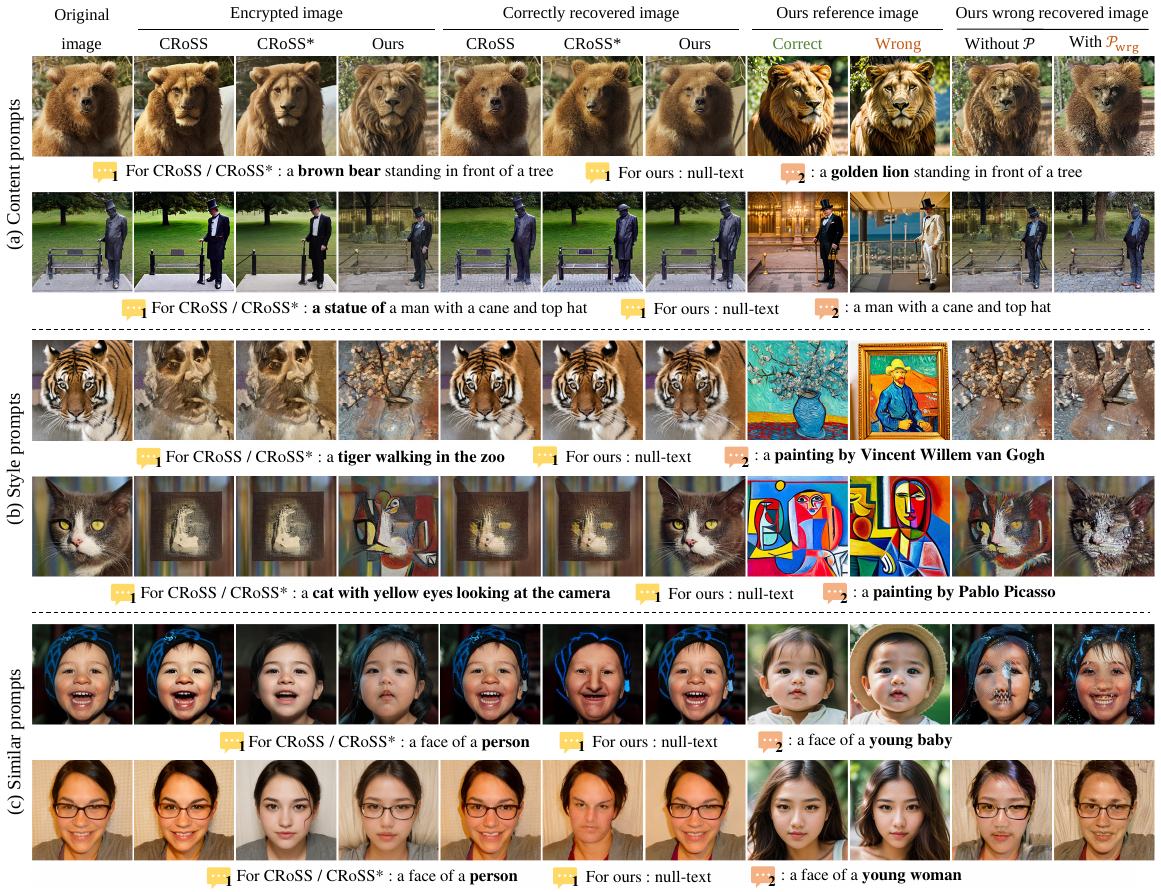}
  \caption{The visual comparison of DiffStega and CRoSS families on UniStega dataset with different prompts. For DiffStega (ours), we categorize the recovery it into three possible scenarios. Besides recovering with correct private key, malicious attackers may attempt recovery without any password, ignoring the use of Guidance Injection and Noise Flip, or recovery with the wrong password \textcolor[rgb]{0.773, 0.353, 0.067}{$\mathcal{P}_{wrg}$}. Note that although we display prompt 1 bellow images, DiffStega still uses null-text as prompt 1 instead. CRoSS* uses two diffusion models consistent with DiffStega rather than a single model in CRoSS.}
  \label{fig:main_result}
\end{figure*}

\subsection{Security Guarantee}
% 对比cross改善了哪些问题，我们的key的关键作用
% 密码虽然说是随机种子，但其实可以规定一个特别长的随机数，将其分段作为随机种子产生噪声，或者规定多个密码生成随机种子，再进行融合得到总的噪声
Previous work uses prompt 1 as private key and prompt 2 as public key~\cite{yu2024cross}. They believe that attackers can only guess the prompt 1 by exhaustive method, being unable to judge which is the true $I_{ori}$ from candidate recovered images. However, we believe that potential attackers usually conduct many investigations on the target. For example, if the target is a military organization, the private key is likely to be related to weapons and equipment. Moreover, frequently transferring private prompt 1 faces risks. In order to make the steganography and recovery only depend on the specified password, DiffStega encrypts the original image with the guidance of images generated with passwords.

In the whole pipeline, $\mathcal{P}_{crt}$ only needs to be transmitted once and is independent of the original image contents.
% For attackers, the original image cannot be correctly restored without $\mathcal{P}_{crt}$. In contrast, 
For trusted parties, $I_{ref}$ can be losslessly reproduced through $\mathcal{P}_{crt}$ and guides recovery stage to restore the original image. Null-text prompt 1 makes it impossible for an attacker to speculate on the original content of the image, which makes the wrong decryption less likely to be the original content. 
% It should be noted that our method does not require additional training process.
And our pipeline can resist steganalysis and many kinds of distortion because of the inherent advantages of diffusion models, which has been explained in~\cite{yu2024cross}.

\section{Experiment}
\subsection{Implementation Details}
\label{sec:exp_details}
For all experiments in this paper, we use the pre-trained SD v1.5\footnote{\url{https://huggingface.co/runwayml/stable-diffusion-v1-5}} in the forward diffusion of hiding stage and the reverse diffusion of the recovery stage. And we use PicX\_real\footnote{\url{https://huggingface.co/GraydientPlatformAPI/picx-real}} in RefGen. We use SD v1.5 for experiments on style prompts in the reverse diffusion of hiding stage and the forward diffusion of the recovery stage, but PicX\_real for other experiments.
We set $T=50$, and the mixing coefficient of EDICT is 0.93. We use IP-Adapter-plus~\cite{ye2023ip} in Guidance Injection, and its weight factor is 1. The guidance scale of diffusion models is 1. $\eta=0.05$ in Noise Flip.
The diffusion process for ours is executed over steps $[0, \xi T]$.
% facilitating exact inversion and sufficient editing of the image. 
We set $\xi = 0.7$ for experiments on style prompts and $\xi = 0.6$ for other prompts.
DiffStega uses ControlNet~\cite{zhang2023adding} with additional control images in RefGen for all experiments except for style prompts. For additional control images, DiffStega uses semantic segmentations from OneFormer~\cite{jain2023oneformer} as control images in general, but OpenPose bone images for face images. Experiments on style prompts use no control image.
All experiments are conducted on single Nvidia RTX 3090 GPU, requiring no additional training or fine-tuning.

Since there is only one diffusion-based CIS model, we mainly compare DiffStega with CRoSS~\cite{yu2024cross}. For fair comparison, CRoSS* is the modefied version which uses the same two models used in DiffStega, rather than only SD v1.5 in the original CRoSS. And all models use EDICT inversion.
% rather than DDIM inversion. 
Also, CRoSS families use ControlNet with the same control images in the reverse diffusion of hiding stage and the forward diffusion of the recovery stage when DiffStega uses them. For sufficient inversion steps, $\xi$ is set to 1 for CRoSS families. 
% Since CRoSS families requires sufficient inversion steps to satisfy the target text prompt, $\xi$ is set to 1. 
Meaningful description texts are only used as prompt 1, i.e. private keys, for CRoSS families. On the contrary, DiffStega uses null-text as prompt 1. All models use target texts as prompt 2. We also conduct security and robustness experiments compared with HiNet~\cite{jing2021hinet}. 

\begin{table*}[th]
  \centering
  \scriptsize
  \setlength{\tabcolsep}{2.3pt}
    \begin{tabular}{@{}cccccccccccccccccc@{}}
    \toprule
    \multirow{2}{*}{Method}& \multicolumn{5}{c}{Encryption with steganography} & \multicolumn{4}{c}{Recovery using correct private key}                & \multicolumn{4}{c}{Recovery without any password}                  & \multicolumn{4}{c}{Recovery using wrong password}                  \\ \cmidrule(l){2-6} \cmidrule(l){7-10} \cmidrule(l){11-14} \cmidrule(l){15-18}
    & PSNR$\downarrow$ & SSIM$\downarrow$ & LPIPS$\uparrow$ & ID Sim$\downarrow$ & CLIP Score $\uparrow$ & PSNR$\uparrow$  & SSIM$\uparrow$ & LPIPS$\downarrow$ & ID Sim$\uparrow$ & PSNR$\downarrow$ & SSIM$\downarrow$ & LPIPS$\uparrow$ & ID Sim$\downarrow$ & PSNR$\downarrow$ & SSIM$\downarrow$ & LPIPS$\uparrow$ & ID Sim$\downarrow$ \\ \midrule
CRoSS                   & 19.034           & 0.657            & 0.365           & 0.892              & 26.912                & 21.248          & 0.711          & 0.320             & 0.877            & -                & -                & -               & -                  & -                & -                & -               & -                  \\
CRoSS*                  & \textbf{17.139}  & 0.606            & 0.435           & 0.516              & 28.251                & 19.553          & 0.673          & 0.348             & 0.750            & -                & -                & -               & -                  & -                & -                & -               & -                  \\ \midrule
DiffStega               & 18.611           & {\ul 0.590}      & {\ul 0.461}     & {\ul 0.343}        & \textbf{29.446}       & 23.290          & 0.769          & 0.266             & 0.893            & \textbf{18.491}  & \textbf{0.595}   & \textbf{0.457}  & {\ul 0.331}        & \textbf{17.530}  & \textbf{0.540}   & \textbf{0.476}  & \textbf{0.477}     \\
DiffStega$^\dagger$     & {\ul 18.529}     & \textbf{0.586}   & \textbf{0.462}  & \textbf{0.340}     & {\ul 29.203}          & {\ul 23.365}    & {\ul 0.773}    & {\ul 0.261}       & {\ul 0.902}      & {\ul 18.840}     & {\ul 0.614}      & {\ul 0.429}     & \textbf{0.315}     & {\ul 17.739}     & {\ul 0.555}      & {\ul 0.457}     & {\ul 0.486}        \\
DiffStega$^\ddagger$    & 19.734           & 0.653            & 0.421           & 0.413              & 29.037                & \textbf{23.920} & \textbf{0.785} & \textbf{0.252}    & \textbf{0.924}   & 20.228           & 0.682            & 0.390           & 0.497              & 20.680           & 0.703            & 0.350           & 0.814              \\ \bottomrule
    \end{tabular}
    \caption{Quantitative assessment and ablation results of encrypted images and recovered images on UniStega dataset. CRoSS* is the revised version of CRoSS for fair comparation, which uses two diffusion models consistent with DiffStega rather than a single model in CRoSS.
    DiffStega$^\dagger$ is the ablation comparation when prompt 1 is not null-text but meaningful text. DiffStega$^\ddagger$ is the version without Noise Flip.}
\label{tab:enc_and_rec}
\end{table*}

\subsection{Datasets and Metrics}
\paragraph{Datasets.}
Since our method supports universal CIS under arbitrary text prompts and can be applied to versatile images, we build a dataset UniStega, consisting of 3 subsets with a total of 100 images under different scenarios: (1) UniStega-Content comprises of 42 images with corresponding prompts and target content prompts. It applies to the most common CIS scenario with similar shape but different content; (2) UniStega-Style comprises of 28 images and prompt pairs. The target prompts refer to artworks in the styles of famous artists. It applies to CIS with higher difficulty to guess what the original images are;
(3) UniStega-Similar comprises of 30 images and analogous prompt pairs. It applies to CIS with similar or overly general prompts.
All images are from public dataset COCO~\cite{lin2014microsoft}, AFHQ~\cite{choi2020stargan}, FFHQ~\cite{ffhq}, CelebA-HQ~\cite{celeba} and Internet, center cropped and resized to $512 \times 512$.
We use BLIP~\cite{li2022blip} to generate description prompts, and Llama2~\cite{touvron2023llama} to generate target content prompts with semantic modifications, or artificial adjustment to generate other prompts, following CRoSS~\cite{yu2024cross}.
% See ~\Cref{sec:dataset} for more details.

\paragraph{Metrics.}
We use PSNR, SSIM~\cite{wang2004image}, LPIPS~\cite{zhang2018unreasonable}, and ID Cosine Similarity from Facenet~\cite{schroff2015facenet} for face images to assess the quality of hiding and recovery. We use CLIP Score~\cite{radford2021learning} to evaluate whether the encrypted image matches the target text promt or not. We use NIQE~\cite{mittal2012making} to blindly assess the naturalness of encrypted images.

% \begin{table}[t]
% \centering
%   \scriptsize
%   \setlength{\tabcolsep}{8pt}
%     \begin{tabular}{@{}cccccc@{}}
%     \toprule
%     Method               & PSNR$\downarrow$ & SSIM$\downarrow$ & LPIPS$\uparrow$ & ID Sim$\downarrow$ & CLIP Score $\uparrow$ \\ \midrule
%     CRoSS                & 19.034           & 0.657            & 0.365           & 0.892              & 26.912                \\
%     CRoSS*               & \textbf{17.139}  & 0.606            & 0.435           & 0.516              & 28.251                \\ \midrule
%     DiffStega            & 18.611           & {\ul 0.590}      & {\ul 0.461}     & {\ul 0.343}        & \textbf{29.446}       \\
%     DiffStega$^\dagger$  & {\ul 18.529}     & \textbf{0.586}   & \textbf{0.462}  & \textbf{0.340}     & {\ul 29.203}          \\
%     DiffStega$^\ddagger$ & 19.734           & 0.653            & 0.421           & 0.413              & 29.037                \\ \bottomrule
%     \end{tabular}
%     \caption{Quantitative comparison and ablation results of encrypted images on UniStega dataset.
%     CRoSS* is the revised version of CRoSS for fair comparation, which uses two diffusion models consistent with DiffStega.
%     DiffStega$^\dagger$ sets prompt 1 to not null-text but meaningful text. DiffStega$^\ddagger$ is the version without Noise Flip.}
% \label{tab:enc}
% \end{table}

\subsection{Experimental Results}
\paragraph{Qualitative Results.}
\Cref{fig:main_result} shows the visual comparision of DiffStega and CRoSS families on UniStega dataset. We categorize it into three possible scenarios. The first is that authenticated users recover the original images with correct private key.
Although DiffStega makes more modifications to the original images, it performs more accurate recovery.
The others are that attackers attempt to recover without any password, i.e. not using RefGen and Noise Flip, or recovering with a wrong password.
Attackers even hardly recover the correct category of original objects.
For style prompts, the encrypted image of CRoSS loses too many details of original images, resulting in a significant difference between recovered images and the original. 
For similar prompts, using a single diffusion model, the encrypted images of CRoSS are almost the same as the original. CROSS* fails in recovery because of inversion errors and general prompt 1.
However, with the guidance of reference image and fewer diffusion steps, DiffStega could still achieves satisfactory performance.

\paragraph{Quantitative Results.}
For encryption with steganography shown in~\Cref{tab:enc_and_rec}. The smaller the similarity to the original image, the better the hiding performance. DiffStega makes the encrypted images distinct from the original ones. Due to the lack of more effective guidance, CRoSS has the worst encryption performance. Meanwhile, the CLIP Scores between encrypted images and prompt 2 shows that DiffStega has better consistency with target prompts. Moreover, DiffStega has stronger identity hiding capability of face images than CRoSS families.
For recovery of different scenarios shown in~\Cref{tab:enc_and_rec}. The greater the similarity to the original image, the better the recovery performance with correct private key. For scenarios of wrong recovery, the opposite is true. DiffStega shows better recovery performance than CRoSS families. 

\begin{figure}[t]
  \centering
  \includegraphics[width=0.9\columnwidth]{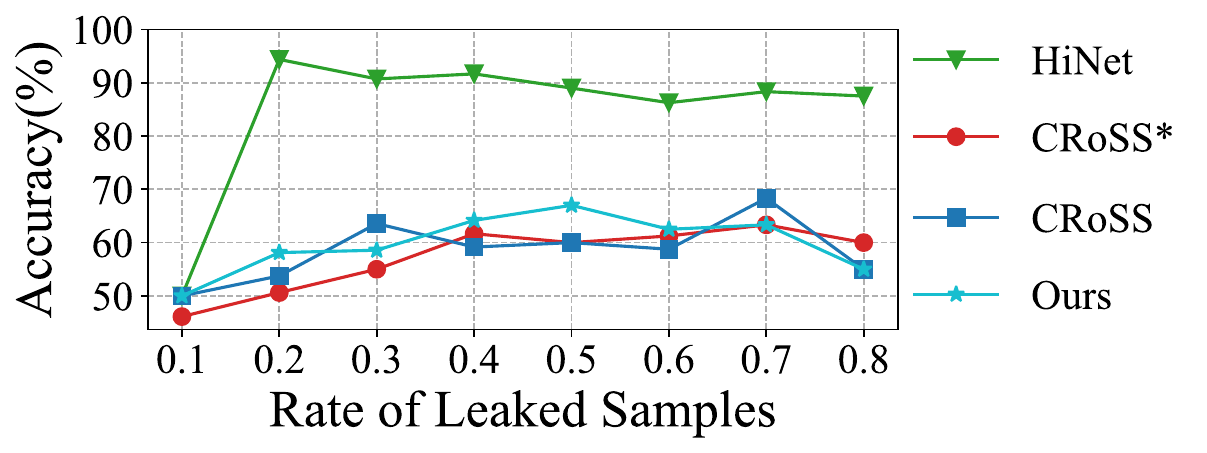}
    \caption{Deep steganalysis accuracy by XuNet. As the rate of leaked samples increases, the closer the curve approximates 50\%, the more secure the method is. Diffusion-based methods are similar.
    }
  \label{fig:deep-ana}
\end{figure}

\begin{figure}[t]
        \small
	\centering
	\begin{subfigure}[b]{0.49\linewidth}
		\centering		\includegraphics[width=\textwidth]{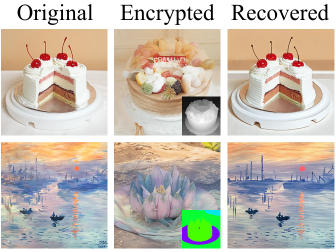}
		\caption{Different controls.}		
	\end{subfigure}
        \hfill
         \begin{subfigure}[b]{0.49\linewidth}
		\centering		\includegraphics[width=\textwidth]{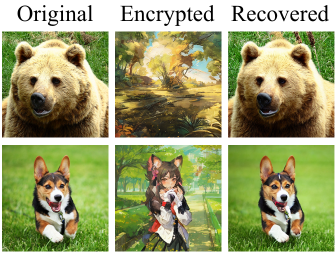}
		\caption{Different checkpoints.}
	\end{subfigure}
	\caption{The encrypted and recovered images of DiffStega with different controls (bottom right) and checkpoints of diffusion models.}
    \label{fig:control_main}
\end{figure}

\begin{figure*}[t]
        \small
	\centering
	\begin{subfigure}[t]{0.27\linewidth}
		\centering		\includegraphics[width=\textwidth]{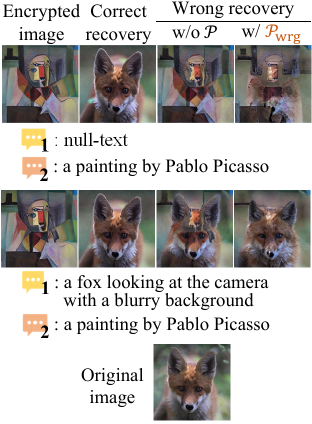}
		\caption{Null-text and meaningful prompt.}
        \label{fig:with_prompt1}	
	\end{subfigure}
        \hfill
         \begin{subfigure}[t]{0.352\linewidth}
		\centering		\includegraphics[width=\textwidth]{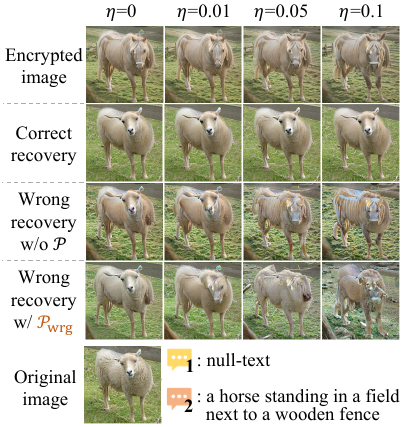}
		\caption{Different Noise Flip scales $\eta$.}\label{fig:ablation_flip}
	\end{subfigure}
        \hfill
	\begin{subfigure}[t]{0.36\linewidth}
		\centering		\includegraphics[width=\textwidth]{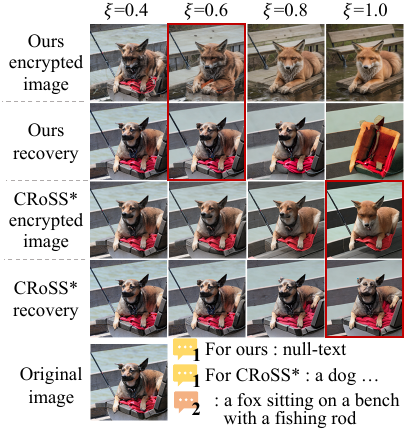}
		\caption{Different diffusion steps with $\xi$.}\label{fig:ablation_xi_main}	
	\end{subfigure}
	\caption{Ablations. (a) The comparation of encrypted and recovered images when prompt 1 is null-text and meaningful text for DiffStega. (b) Visual comparisons of DiffStega with different Noise Flip scales. The first row shows the original image and the encrypted images. (c) The encrypted and recovered images with different diffusion steps according to $\xi$. CRoSS* is the revised version of CRoSS for fair comparation.}
\end{figure*}

\begin{table}[t]
  \centering
  \scriptsize
  \setlength{\tabcolsep}{10pt}
\begin{tabular}{@{}cccccc@{}}
\toprule
                               & HiNet                           & CRoSS & CRoSS* & Ours & Original image \\ \midrule
NIQE $\downarrow$               & \textbf{3.125} & 3.601 & 3.795  & {\ul 3.408}  & 3.083          
% \\ $\mathcal{R}_{anti}\downarrow$ &                                 &       &        &  
\\ \bottomrule
\end{tabular}
\caption{
NIQE scores that indicate the quality of encrypted images. All methods is relatively as natural as the original images. 
}
\label{tab:niqe}
\end{table}

% \begin{figure}[t]
%   \small
%   \centering
%   \includegraphics[width=\linewidth]{images/with_prompt1.pdf}
%   \caption{The comparation of encrypted and recovered images when prompt 1 is null-text and meaningful text for DiffStega.}
%   \label{fig:with_prompt1}
% \end{figure}

% \begin{figure}[t]
% \small
%   \centering
% \includegraphics[width=0.9\linewidth]{images/ablation_flip.pdf}
%   \caption{Visual comparisons of DiffStega with different Noise Flip scales $\eta$. The first row shows the original image and the encrypted images.}
%   \label{fig:ablation_flip}
% \end{figure}

% \begin{figure}[t]
% \small
%   \centering
% \includegraphics[width=0.9\linewidth]{images/ablation_xi_main.pdf}
%   \caption{The encrypted and recovered images with different diffusion steps according to $\xi$. CRoSS* is the revised version of CRoSS for fair comparation.}
%   \label{fig:ablation_xi_main}
% \end{figure}

\paragraph{Imperceptibility and Security.}
\Cref{tab:niqe} shows that ours has similar NIQE score to the original images, hardly suspected by people. For anti-analysis security, We use XuNet~\cite{xu2016structural} to distinguish the encrypted images to general images without steganography. As shown in~\Cref{fig:deep-ana}, the closer the detection accuracy approximates 50\%, the more secure the method is. DiffStega shares the similar performance with CRoSS families, much better than HiNet.

\paragraph{Robustness and Controllability.}
Real-ESRGAN~\cite{wang2021realesrgan} is used to perform nolinear image enhancement for degradations, denoted as Gaussian deblur and JPEG enhancer. As shown in~\Cref{tab:robust_main}, DiffStega has similar robustness to CRoSS, while HiNet suffers significant drops in PSNR. \Cref{fig:control_main} shows that our pipeline is flexible and applicable to different controls and checkpoints\footnote{\url{https://huggingface.co/stablediffusionapi/majicmix-fantasy}}.
% , bring better imperceptibility while maintaining satisfying recovery.

\subsection{Ablation Study}
\paragraph{Influence of Null-text Prompt.}
As shown in~\Cref{fig:with_prompt1}, meaningful prompt 1 narrows the similarity between the wrong decrypted image and the original image.
Considering that publicly disclosing meaningful prompt 1 is equivalent to letting everyone know what is encrypted, but only brings minimal performance improvement as shown in~\Cref{tab:enc_and_rec}, we prefer to set null-text as prompt 1 for better security.

\paragraph{Influence of Noise Flip.}
As shown in~\Cref{tab:enc_and_rec} and~\Cref{tab:enc_and_rec} for DiffStega$^\ddagger$, Noise Flip increases the reliance on the private password. Without Noise Flip, DiffStega has slight improvement in the quality of recovery, but double performance drop on encryption and wrong decryption.
As $\eta$ grows in~\Cref{fig:ablation_flip}, although wrong decrypted images are getting further away from the original, the quality of the encrypted images is getting worse and worse. Therefore $\eta=0.05$ is sufficient. The second row demonstrates that different Noise Flip scales have limited influence of the correct recovery.

\paragraph{Influence of $\xi$.}
As shown in~\Cref{fig:ablation_xi_main}, DiffStega achieves distinct change and meets the target description when $\xi \geq 0.6$. However, only when $\xi$ is close to 1 can CRoSS families make the encrypted image distinct from the original. The red box indicates the values we recommended in the experiments.

\section{Limitation}
Because there is inevitable error in diffusion inversion methods, DiffStega suffers the poor recoverability if $\xi$ is too large.
Meanwhile, DiffStega introduces additional processes of generating reference images, which means more computational overhead. However, the additional cost will gradually be negligible with fast sampling methods~\cite{luo2023latent}.

\begin{table}[t]
  \centering
  \scriptsize
  \setlength{\tabcolsep}{1.2pt}
\begin{tabular}{cccccc}
\toprule
Method    & Clean  & Gaussian blur   & Gaussian deblur & JPEG compression & JPEG enhancer \\ \midrule
HiNet     & 46.152 & 10.262          & 10.992          & 10.946           & 10.856           \\
CRoSS     & 21.248 & {\ul 20.080}    & \textbf{19.354} & {\ul 20.195}     & {\ul 19.267}     \\
CRoSS*    & 19.553 & 16.371          & 16.696          & 16.697           & 16.032           \\
DiffStega & 23.290 & \textbf{20.849} & {\ul 18.620}    & \textbf{21.161}  & \textbf{20.154} \\ \bottomrule
\end{tabular}
    \caption{
    PSNR(dB) results of recovered images on UniStega dataset when encrypted images suffer degradations.
    CRoSS* uses two diffusion models consistent with DiffStega.
    % CRoSS* is the revised version of CRoSS for fair comparation.
    For Gaussian blur and deblur, the kernel size is 7 and $\sigma$ is 10 . $Q$ is 40 for JPEG  degradations.}
\label{tab:robust_main}
\end{table}

\section{Conclusion}
This paper proposes DiffStega, cleverly designed for general coverless image steganography with diffusion models.
We uses the pre-determined password as private key, and propose Noise Flip to achieve high quality steganography and nearly undistorted recovery of the original images.
Extensive experiments show our superiority compared with previous methods.
How to directly influence the image generation process with passwords is a promising research topic in the future.

\section*{Acknowledgments}
This work was supported in part by the China Postdoctoral Science Foundation under Grant Number 2023TQ0212 and 2023M742298, in part by the Postdoctoral Fellowship Program of CPSF under Grant Number GZC20231618, in part by the National Natural Science Foundation of China under Grant 62301310 and 62301316, and in part by the Shanghai Pujiang Program under Grant 22PJ1406800.

\section*{Contribution Statement}
Yiwei Yang and Zheyuan Liu contributed equally.
% to this work.

\bibliographystyle{named}
\bibliography{arxiv}

\end{document}